\documentclass[conference]{IEEEtran}  

\usepackage{xparse} 
\usepackage{amsmath} 
\usepackage{amssymb}
\usepackage{amsthm}
\usepackage{amsfonts} 
\usepackage{color}

\usepackage{graphicx}
\usepackage{subfig}
\graphicspath{{Figures/}}

\makeatletter
\long\def\@makecaption#1#2{%
  \vskip\abovecaptionskip
  \sbox\@tempboxa{{\small\bfseries #1.} #2}%
  \ifdim \wd\@tempboxa >\hsize
    {\footnotesize\bfseries #1.} #2\par
  \else
    \global \@minipagefalse
    \hb@xt@\hsize{\hfil\box\@tempboxa\hfil}%
  \fi
  \vskip\belowcaptionskip}
\makeatother

\usepackage[ruled,vlined]{algorithm2e}
\usepackage{etoolbox}%
\providetoggle{beginFlag}
\settoggle{beginFlag}{true}

\usepackage{hyperref}

\usepackage{cleveref}
\crefformat{equation}{(#2#1#3)}
\Crefname{figure}{Fig.}{Figs.}
\crefname{figure}{Fig.}{Figs.}
\crefformat{figure}{#2Fig.~#1#3}
\Crefformat{figure}{#2Fig.~#1#3}
\crefformat{table}{#2Table~#1#3}
\Crefformat{table}{#2Table~#1#3}
\crefformat{section}{#2Sec.~#1#3}
\Crefformat{section}{#2Sec.~#1#3}
\crefformat{problem}{#2Problem~#1#3}
\Crefformat{problem}{#2Problem~#1#3}
\crefname{appsec}{Appendix}{Appendices}

\usepackage{url}
\usepackage{graphicx}

\usepackage{array}
\usepackage{multirow}
\usepackage{float}
\usepackage{mathrsfs}
\usepackage{listings}
\usepackage{rotating}
\usepackage{overpic}
\usepackage{tikz}
\usetikzlibrary{arrows.meta, positioning, fit}

\usepackage{booktabs}

\usepackage{rigidnotation}
\SetConciseNotation
\usepackage{xparse}
\usepackage{tikz}

\newcommand{\lrss}[5]{%
	\setbox1=\hbox{\ensuremath{^{#1}}}%
	\setbox2=\hbox{\ensuremath{_{#2}}}%
	\setbox5=\hbox{\ensuremath{#5}}%
	\setbox6=\hbox{\ensuremath{^{#1#3}}}%
	\setbox7=\hbox{\ensuremath{_{#2#4}}}%
	\setbox8=\hbox{\ensuremath{^{#3}}}%
	\setbox9=\hbox{\ensuremath{_{#4}}}%
	\hspace{\ifnum\wd1>\wd2\wd1\else\wd2\fi}%
	\ensuremath{\copy5%
		^{\hspace{-\wd1}\hspace{\wd1}\hspace{\wd8}%
			\hspace{-\wd6}\hspace{-\wd5}#1\hspace{\wd5}#3}%
		_{\hspace{-\wd2}\hspace{\wd2}\hspace{\wd9}%
			\hspace{-\wd7}\hspace{-\wd5}#2\hspace{\wd5}#4}%
}}

\NewDocumentCommand\bbm{}{ \begin{bmatrix} }
\NewDocumentCommand\ebm{}{ \end{bmatrix} }
\NewDocumentCommand\Vector{m}{ \boldsymbol{\mathbf{#1}} }

\NewDocumentCommand\Norm{m}{\left\Vert#1\right\Vert }

\NewDocumentCommand\Real{}{ \mathbb{R} }
\NewDocumentCommand\NonnegativeReal{}{ \Real_+ }

\NewDocumentCommand\AbsoluteValue{m}{ \left\vert#1\right\vert }
\NewDocumentCommand\Abs{m}{ \AbsoluteValue{#1} }

\NewDocumentCommand\StableSet{}{ \mathcal{S} } 
\NewDocumentCommand\UnstableSet{}{ \overline{\mathcal{S}} } 
\NewDocumentCommand\PenetratingSet{}{ \mathcal{P} } 
\NewDocumentCommand\PenetrationFreeSet{}{ \overline{\mathcal{P}} }
\NewDocumentCommand\AllPlacementsSet{}{\mathcal{A}}

\IEEEoverridecommandlockouts

\title{\LARGE \bf Generating Stable Placements via Physics-guided Diffusion Models}

\author{Philippe Nadeau$^1{}^*$, Miguel Rogel$^1{}^*$, Ivan Bilić$^2{}^*$, Ivan Petrović$^2{}$, and Jonathan Kelly$^1{}^\ddagger$
\thanks{$^*$Equal contribution.}
\thanks{$^1$STARS Laboratory, University of Toronto Institute for Aerospace Studies, Toronto, Ontario, Canada. {\tt\footnotesize <firstname>.<lastname>@robotics.utias.utoronto.ca}}
\thanks{$^2$Laboratory for Autonomous Systems and Mobile Robotics, University of Zagreb Faculty of Electrical Engineering and Computing, Zagreb, Croatia. {\tt\footnotesize <firstname>.<lastname>@fer.hr}}
\thanks{$^\ddagger$Jonathan Kelly is a Vector Institute Faculty Affiliate. This research was supported in part by the Canada Research Chairs program.}}

\begin{document} 

\maketitle 
\thispagestyle{empty}
\pagestyle{empty}

\begin{abstract}
Stably placing an object in a multi-object scene is a fundamental challenge in robotic manipulation, as placements must be penetration-free, establish precise surface contact, and result in a force equilibrium.
To assess stability, existing methods rely on running a simulation engine or resort to heuristic, appearance-based assessments.
In contrast, our approach integrates stability directly into the sampling process of a diffusion model.
To this end, we query an offline sampling-based planner to gather multi-modal placement labels and train a diffusion model to generate stable placements.
The diffusion model is conditioned on scene and object point clouds, and serves as a geometry-aware prior.
We leverage the compositional nature of score-based generative models to combine this learned prior with a stability-aware loss, thereby increasing the likelihood of sampling from regions of high stability.
Importantly, this strategy requires no additional re-training or fine-tuning, and can be directly applied to off-the-shelf models.
We evaluate our method on four benchmark scenes where stability can be accurately computed. Our physics-guided models achieve placements that are 56\% more robust to forceful perturbations while reducing runtime by 47\% compared to a state-of-the-art geometric method.

\end{abstract}

\section{Introduction}
\label{sec:introduction}

The task of planning to stably place an object among other objects, henceforth referred to as \textit{stable placement planning}, is a core element of several higher level tasks, like construction \cite{wermelinger_2021_grasping, johns_2023_framework}, scene rearrangement \cite{srinivas_2023_busboy}, and dense packing \cite{wang_2021_dense}.
The challenge lies in satisfying strict geometric and physical constraints that characterize the problem.
A placement pose must establish precise surface contact with objects in the scene while avoiding penetration with any other object. At the same time, it must result in a force equilibrium to prevent toppling. 
Ideally, a placement planner would sample directly from the space of valid poses; however, defining this subset is complex.
Only a very small proportion of all poses in the workspace result in valid placements, making random sampling or searching approaches highly inefficient \cite{nadeau_2025_planning}.

We argue that a principled approach to determine stable placements requires: (i) a geometry-aware placement algorithm, (ii) reasoning about scene equilibrium, and (iii) a stability verifier to assess the outcome. To reason about scene equilibrium, existing methods either rely on executing a simulation engine \cite{simeonov_2023_shelving, yoneda_2023_stabilityfield}, or assess placement quality solely based on appearance \cite{newbury_2021_learning, paxton_2022_predicting}. The former is time-consuming, while the latter requires making assumptions about object inertial parameters.
\begin{figure}[t]
    \centering
    \setlength{\fboxsep}{0pt}%
    \setlength{\fboxrule}{0.5pt}%
    \fbox{\includegraphics[width=1\linewidth]{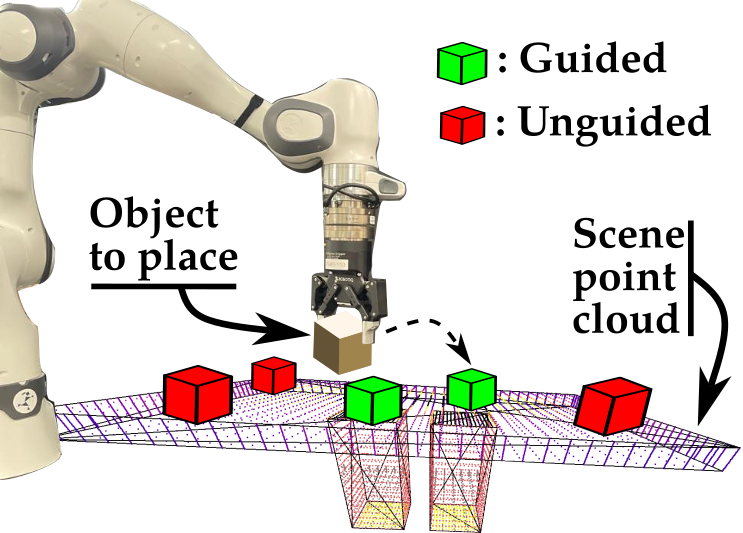}}
    \vspace{-5mm}
    \caption{Our proposed method consists of a learned diffusion-based prior that takes a processed scene point cloud (here, a slab supported by two legs) and generates stable placements. We show that our proposed guidance scheme enables robots to find more stable placement poses faster.}
    \label{fig:frontpage}
\end{figure}
We propose a simulation-free approach that addresses stability in a principled manner, by merging requirements (i) and (ii): enhancing the placement algorithm with a physics-based feature, obtained from \cite{nadeau_2025_robustness}, termed \textit{robustness}. Robustness describes the maximum force that can be applied at a point in a scene before any object moves.

Due to the lack of publicly available datasets that capture the multi-object stability information necessary for our approach, we employ a state-of-the-art sampling-based planner \cite{nadeau_2025_planning} to act as an expert. We construct four scenes, and use the planner offline to generate multi-modal stable placement examples and physics-based robustness features. Our scenes comprise flat placement surfaces, similar to recent works which consider stability \cite{simeonov_2023_shelving, yoneda_2023_stabilityfield, newbury_2021_learning, paxton_2022_predicting}. While such scenes do not capture the full complexity of the real world, they allow us to precisely assess the quality of stability reasoning and a certain degree of generalization to unseen data. 

Diffusion models are capable of capturing complex distributions and excel at conditional and controllable generation \cite{song_2023_lgd, janner_2022_planning}.
Therefore, we introduce a physics-guided score-based generative model for stable placement planning. The compositional nature of score-based generative models allows us to combine multiple likelihood terms—such as the prior and differentiable loss—into a single sampling distribution.
Our model is trained to generate stable placements conditioned only on scene and object geometry, serving as a prior with no explicit physics information. During sampling, we introduce a stability-aware diffusion guidance to inform the learned prior about scene statics, steering the sampling process toward regions of high stability, as shown in \cref{fig:frontpage}.
Our guidance is easy to compute and does not affect model training, making it applicable to off-the-shelf models. Compared to our geometry-based learned prior and \cite{nadeau_2025_planning}, we observe our guidance can increase the placement validity rate up to $40\%$ on unseen scenes, and overall scene robustness after placing by $56\%$.
We also discuss architectural details that we identify as crucial for enabling the model to generalize, independent of guidance.
To the best of our knowledge, our work is the first to incorporate a physics-informed prior into a generative stable placement model.

\section{Related Work}
\label{sec:related_work}
\subsection{Algorithms for Object Placement Planning}
\label{sec:rel_work_learning_free}
Stable object placement planning has typically been tackled in a search-and-evaluate fashion, where a grid-search over the space of poses is performed to produce candidate placements, and a validation step is executed to reject unfeasible poses.
In cluttered or complex scenes, a very small proportion of poses results in stable placements \cite{nadeau_2025_planning} and many methods rely on hand-crafted heuristics to accelerate the search.
For instance, \cite{wang_2021_dense} performs a grid-search over the space of poses and score candidate placement poses according to a heightmap minimization heuristic criterion. The stability of best-scoring candidate poses is then evaluated until a stable placement is found.
In contrast, our approach samples placement poses more densely in regions where a stable outcome is more likely.

To reduce the size of the search space, some methods have focused on placing objects on horizontal surfaces.
\cite{harada_2014_validating} matches sets of surface planes on the object to a horizontal plane in the environment and relies on rejection sampling to find a valid placement pose.
Similarly, \cite{haustein_2019_object} projects the centre of mass of the object onto its convex hull to determine contact points that could support the object when placed on a horizontal surface.
\cite{baumgartl_2014_fast} also relies on such a projection, and defines a GPU-accelerated geometric approach to local placement pose optimization, relying on a heuristic for pose evaluation.
While our approach is also evaluated in scenes with horizontal surface planes, we do not rely on any operation that would only work in such scenes.

For an object placement planning algorithm to be applicable in a broad range of environments, simple stability evaluation heuristics might not suffice.
In this context, \cite{chen_2021_planning} proposes a system that considers a large number of rigid object structures in a hierarchical fashion, and evaluates the stability of each structure by applying random wrench disturbances to the object via an optimization-based method proposed by \cite{maeda_2009_new}.
Meanwhile, \cite{lee_2023_object} proposes to embed a dynamics simulator within a task planning framework to easily evaluate the stability of proposed placements, highlighting long-lasting run times.
In contrast to these methods, our proposed approach does not require the use of dynamics simulators or applying random wrenches on candidate placements poses.

To improve stable placement planning efficiency, \cite{nadeau_2025_planning} proposes to consider contact points instead of poses when placing objects, making use of the assembly robustness assessment in \cite{nadeau_2025_robustness} to sample contact points in the environment that are more likely to result in an stable scene.
Although this work makes use of the robustness assessment algorithm defined in \cite{nadeau_2025_robustness}, our approach deals directly with poses and can result in faster planning time when compared to \cite{nadeau_2025_planning}.

\subsection{Learning-based Placement Planning}
\label{sec:rel_work_learning_based}
The great computational complexity of the object placement planning problem has motivated the development of learning-based methods that can improve placement planning speed.
\cite{jiang_2012_learning} proposes a learning algorithm taking hand-crafted features (i.e., supporting contacts, gravity caging, object shape) into consideration to estimate the quality of placements in terms of stability and semantic preference, but only considers serial kinematic object assemblies.
\cite{newbury_2021_learning} divides the overall problem of stably placing an object on a horizontal surface into placement generation and quality evaluation, and train two separate neural networks to tackle each sub-problem.
In the same vein, \cite{paxton_2022_predicting} relies on a trained scene discriminator to score the realism of a potential placement and on a relational classifier to estimate the likelihood that a pair of objects satisfy some semantic relationship.

Acknowledging the typical requirement that objects should not move after a placement took place, 
\cite{simeonov_2023_shelving, yoneda_2023_stabilityfield} propose neural network architectures capable of generating candidate placement poses (i.e., diffusion models), but require the use of off-the-shelf dynamics simulators. In contrast to \cite{simeonov_2023_shelving, yoneda_2023_stabilityfield}, we do not require the use of a dynamics simulator in the placement generation process.

In \cite{yoneda_2023_stabilityfield} an architecture based on \cite{ho_2020_ddpm} and \cite{wang_2019_dgcnn} is proposed to infer stable placements of rigid objects with a diffusion model. Given a scene and an object to be placed, a diffusion process iteratively perturbs the pose of the object such that it ends up being close to stable. A dynamics simulator is then used to let the object settle, with the displacement incurred at this stage being used for evaluation purposes and compared to a random sampling baseline.

In \cite{simeonov_2023_shelving}, three types of scenes are considered and a model based on \cite{ho_2020_ddpm} is trained for each environment with the objective of generating placements that are close to a pre-defined (or \textit{correct}) orientation. Like \cite{yoneda_2023_stabilityfield} and this work, \cite{simeonov_2023_shelving} performs iterative pose denoising conditioned on scene and object points clouds to generate placements. In comparison with \cite{simeonov_2023_shelving}, our work explicitly considers placement stability and requires that generated placements be non-penetrating and result in a force equilibrium in the scene.

\section{Background}
\label{sec:background}

\subsection{Stable Object Placement Planning}
\label{sec:placement_planning_problem_def}
Given a scene of $N$ objects in frictional contact whose pose, shape, mass distributions, and coefficients of friction are known, the stable object placement planning problem consists of determining a pose for an object to be placed such that the following two criteria are met:
\begin{enumerate}
    \item (Penetration-free): No pair of objects shall have an overlapping volume.
    \item (Stable): The sum of all forces and torques on any object shall be zero.
\end{enumerate}
A pose for which the two criteria above are met is considered \textit{valid}, any other pose is considered \textit{invalid}.

\subsection{Assembly Robustness}
\label{sec:robustness_def}
This work makes use of the \textit{assembly robustness}, defined in \cite{nadeau_2025_robustness}, that expresses the capacity of an assembly of rigid objects in frictional contact to withstand external forces without moving.
Robustness is formally defined as
\begin{equation}
    \label{eq:rob_def}
    \text{R} : \Real^3 \times \mathbb{S}^2 \rightarrow \NonnegativeReal,
\end{equation}
mapping a pair $(\Pos \in \Real^3~,~\Vector{e} \in \mathbb{S}^2)$ to the maximum force magnitude $\text{R} \in \NonnegativeReal$ that can be exerted on point $\Pos$ in the direction $\Vector{e}$ before any motion occurs, with $\mathbb{S}^2$, the unit sphere.
Computing the robustness of an assembly to an external force involves solving for reaction forces, and considering multi-object slipping and toppling phenomena.
We posit that the robustness information can be a useful intermediate representation and provide a way to encode complex statics in point-cloud features.
In this work, we use the algorithm defined in \cite{nadeau_2025_robustness} to compute the robustness of scene objects to normal forces applied at surface points, which we simply refer to as the robustness.

As highlighted in \cite{nadeau_2025_planning}, the robustness statistics (e.g., \textit{minimum} or \textit{median} robustness), computed across a scene for a given placement pose, can be used to quantitatively compare between placements and formally evaluate placement quality from a stability standpoint.

\subsection{Diffusion Models}
\label{sec:diffusion_models_def}
This work employs diffusion models  \cite{ho_2020_ddpm, song_2019_generative} to model the multimodal distribution of stable placement poses. Diffusion models approximate the sampling process from a distribution $p(\mathbf{x}_0)$, by marginalizing over a Markov chain of latent variables $\mathbf{x}_{1:T}$ with
\begin{align}
    p(\mathbf{x}_0) = \int p(\mathbf{x}_{T)}\prod_{t=1}^{T}p(\mathbf{x}_{t-1} | \mathbf{x}_{t}) d\mathbf{x}_{1:T}.
\end{align}
At each step of the Markov chain, additive noise  defined by noise schedule $\{\alpha_{t}\}_{i=1}^T$ is mixed with the latent variable $\mathbf{x}_t$ according to the noising process
\begin{equation} 
  \mathbf{x}_t = \sqrt{\alpha_t}\mathbf{x}_{t-1} + \sqrt{1 - \alpha_t}\boldsymbol{\epsilon}_{t-1}, \quad \boldsymbol{\epsilon}_{t-1}\sim \mathcal{N}(\mathbf{0}, \mathbf{I}).
\end{equation}
Thus, as $T\to\infty$, the original sample $\mathbf{x}_0$ becomes indistinguishable from Gaussian noise, such that $p(\mathbf{x}_{T})$ can be defined as a unit Gaussian prior. The purpose of diffusion models is to learn the mapping from $\mathbf{x}_T\sim\mathcal{N}(\mathbf{0};\mathbf{I})$ back to $\mathbf{x}_0 \sim p(\mathbf{x}_0)$, which can be achieved by minimizing
the training objective
\begin{align}
    \label{eq:loss_function}
    \mathcal{L}(\mathbf{x}_0, \boldsymbol{\theta})=\sum_{\mathbf{x}_0\in \mathcal{X}} \sum_{t=1}^T \|\hat{\mathbf{x}}_{\boldsymbol{\theta}}(\mathbf{x}_t,t)-\mathbf{x}_0\|^2,
\end{align}
where the model $\hat{\mathbf{x}}_{\boldsymbol{\theta}}$ parameterized by $\boldsymbol{\theta}$ learns to predict the original data point $\mathbf{x}_0$, sampled from dataset $\mathcal{X}$, from a noisy sample $\mathbf{x}_t$.

In order to generate a new sample $\hat{\mathbf{x}}_0$, noise is iteratively removed according to an inverse Langevin dynamics sampling process 
\begin{align}
    \label{eq:langevin_dynamics}
    \mathbf{x}_{t-1} = \beta_t(\mathbf{x}_t + \eta_t\hat{\mathbf{x}}_{\boldsymbol{\theta}}(\mathbf{x}_t,t)) + \boldsymbol{\xi}_t, \quad \boldsymbol{\xi}_t \sim \mathcal{N}(\mathbf{0};\sigma_t \mathbf{I}),
\end{align}
starting from a Gaussian noise sample $\mathbf{x}_T\sim\mathcal{N}(\mathbf{0};\mathbf{I})$ \cite{song_2019_generative}. Here, variables $\beta_t$, $\eta_t$ and $\sigma_t$ are obtained from a scheduler which regulates the variance level at each sampling step.
Due to the score-based formulation of diffusion models \cite{song_2019_generative}, the sampling process in \cref{eq:langevin_dynamics} can be additionally guided by a differentiable cost function $\mathcal{J}:\mathbb{R}^n\to\mathbb{R}$ taking noisy data samples $\mathbf{x}_t$ as input. Let the probability distribution $q(\mathbf{x}_t)\propto \exp{-\mathcal{J}(\mathbf{x}_t})$ assign high probabilities to realizations of $\mathbf{x}_t$ which incur a low cost. The gradient of the log-likelihood of $q(\mathbf{x_t})$ can be incorporated into the sampling process as
\begin{align}
    \label{eq:langevin_guidance}
    \mathbf{x}_{t-1} = \beta_t(\mathbf{x}_t + \eta_t(\hat{\mathbf{x}}_{\boldsymbol{\theta}}(\mathbf{x}_t,t) - \gamma_t\nabla\mathcal{J}(\mathbf{x}_t)) + \boldsymbol{\xi}_t,
\end{align}
where, $\gamma_t \in \mathbb{R}^+$ modulates the relative weights of the gradients in data space. This modified sampling process allows the composition of two objectives in order to generate a sample $\hat{\mathbf{x}}_0$ that has a high likelihood given the observed data $\mathcal{X}$ and which also incurs a low cost under cost function $\mathcal{J}$ \cite{song_2023_lgd}.

\section{Proposed Method}
\label{sec:proposed_method}
We define the problem of finding stable object placements as one of modelling a distribution of placement poses $p(\mathbf{T|\mathbf{P}})$, where $\mathbf{T} \in \rm{SE}(3)$ and $\mathbf{P}$ is an observation of the scene and object. We approximate sampling from $p(\mathbf{T|\mathbf{P}})$ by training a diffusion model to denoise noisy poses from dataset $\mathcal{T}$, described next. In our notation, the subscript $t$ indicates a variable used at the current diffusion step.

\subsection{Dataset Generation}
\label{sec:dataset_generation}
\begin{figure*}[ht]
    \centering
    \subfloat[Table]{
      \includegraphics[width=0.48\columnwidth]{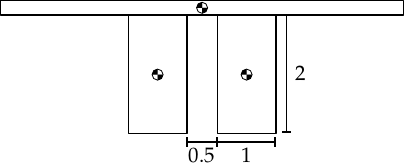}
    }%
    \subfloat[Cantilever]{
      \includegraphics[width=0.48\columnwidth]{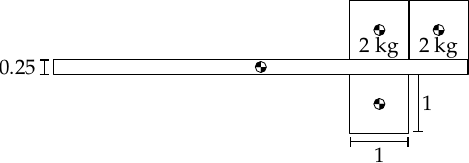}
    }%
    \subfloat[Balance]{
      \includegraphics[width=0.48\columnwidth]{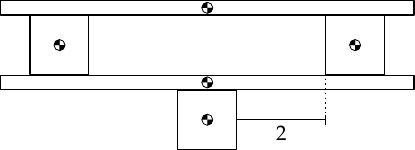}
    }%
    \subfloat[Shelf]{
      \includegraphics[width=0.48\columnwidth]{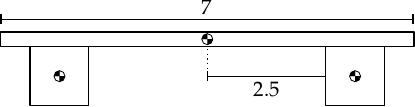}
    }
    \caption{Schematics of the four scenes considered in this work. All distances are in meters and, when not specified, object masses are 1 kg. Objects with identical shapes in the schematics have the same dimensions. The mass distribution of all objects is homogeneous and there is no glue between objects.}
    \label{fig:schematics}
\end{figure*}

We query a placement planner \cite{nadeau_2025_planning} to sample stable placements in the four different scenes whose schematics are shown in \cref{fig:schematics}.
Our scenes are designed such that selecting stable placements requires reasoning about stability.
While \textit{Shelf} affords placing on a large longitudinal section of the top slab, \textit{Table} can topple much more easily if an object is placed too far from the tabletop centre. The \textit{Balance} scene, which is composed of two slabs supported by three legs, requires taking into account, not only the immediate supporting legs, but the whole structure. Finally, our \textit{Cantilever} scene has a large slab on which very little weight can be exerted before toppling happens.

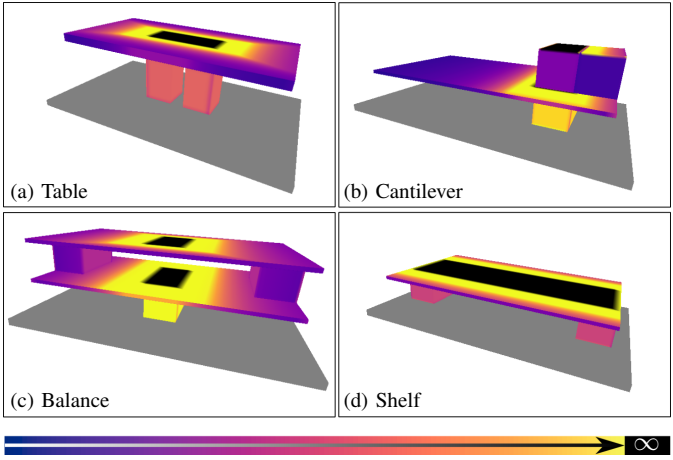
\begin{figure}[t]
    \centering
    \begin{overpic}[width=1\linewidth]{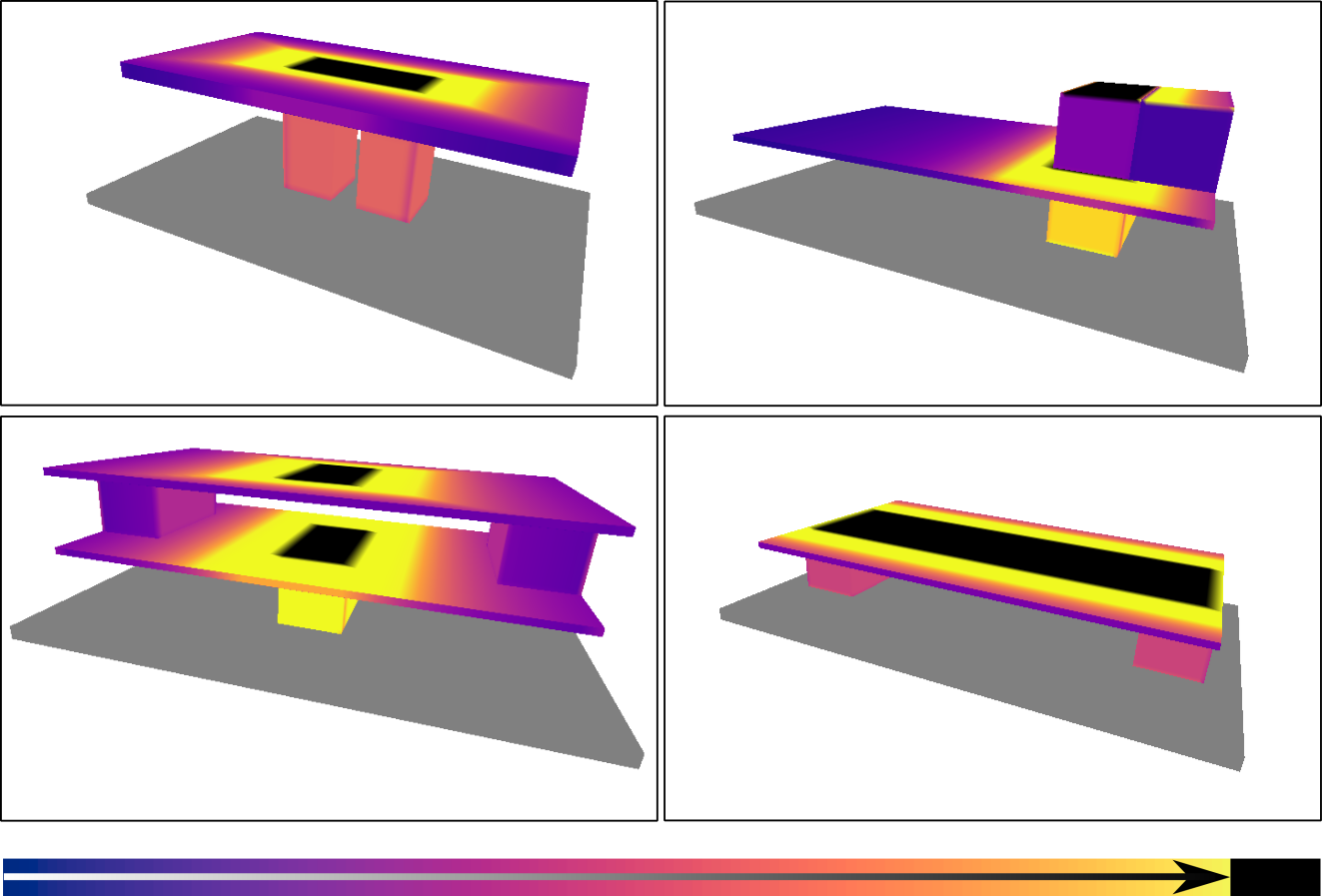}
        \put(1, 38.5){\footnotesize (a) Table}
        \put(51, 38.5){\footnotesize (b) Cantilever}
        \put(1, 7){\footnotesize (c) Balance}
        \put(51, 7){\footnotesize (d) Shelf}
        \put(94.5, 0.6){\color{white}$\infty$}
    \end{overpic}
    \vspace{-5mm}
    \caption{Every surface point is mapped to a robustness value, according to \cite{nadeau_2025_robustness}, color-coded such that lighter colors correspond to higher values and black indicates infinity. We provide robustness values to our model as additional scene point features to enable reasoning about statics.}
    \label{fig:scene_robustness}
\end{figure}

Point clouds comprising 1,024 points are generated for every scene and for the object being placed.
Furthermore, the robustness of each scene point is assessed and made available as an additional feature.
As shown in \cref{fig:scene_robustness}, robustness values can provide a strong prior on sections of the scene that are most useful for stable placement planning \cite{nadeau_2025_planning}, but can also mislead the planner into focusing on regions where the object cannot fit (e.g., on the bottom slab of \textit{Balance}).

For each scene, 800 valid placements are generated with \cite{nadeau_2025_planning} and used for training. 
Each placement is represented as a 9D vector composed of a position vector and a 6D rotation representation \cite{zhou_2019_continuity}.
To increase the size of our dataset and to avoid permitting the model to identify the scene without making use of all information, we perform a sequence of transformations on every input scene point cloud and placement.
A rotation of the scene about the gravitational axis is performed, followed by a general translation and a shuffle of the points.
The parameters of the transformations are randomly selected and are different for every point cloud.
Hence, when testing our method, the scene point clouds used as inputs to the model are guaranteed to have never been seen.

\subsection{Observation Processing}
\label{sec:obs_processing}
We represent observations $\mathbf{P}$ as point clouds composed of scene points $\mathbf{S}$ with cardinality $\Abs{\mathbf{S}}$  and placement object points $\mathbf{O}$ with cardinality $\Abs{\mathbf{O}}$. Each scene point $S_i \in \mathbb{R}^9$ contains its position in space $\mathbf{p}_{S_i}$, a normal vector $\mathbf{n}_{S_i}$ to the local surface on which the point is located, a one-hot encoding representing that the point belongs to the scene cloud, and a robustness feature $\text{R}(S_i)$. Each placement object point $O_i \in \mathbb{R}^8$ contains its position $\mathbf{p}_{O_i}$, normal vector $\mathbf{n}_{O_i}$ and a one-hot encoding representing that the point belongs to the placement object cloud. All vectors are defined with respect to a common reference frame.

Our experiments highlight several architectural choices that are critical to model performance. Specifically, we find that step-level condition updating, local context extraction, and translation equivariance are essential; without them, the model struggles to generate precise and penetration-free placements on unseen data. 

\begin{itemize}
\item \textit{Condition updating.} At each step of the diffusion process, we apply the current noisy pose sample $\hat{\mathbf{T}}_t$ to the object points $\mathbf{O}$, yielding $\hat{\mathbf{O}}_t = \hat{\mathbf{T}}_t \mathbf{O}$. This enables the model to correlate its intermediate predictions with the desired outcome, learning that stable placements are more likely when the object and scene are in close contact.
\item \textit{Local context.} Placement validity is determined by object-to-scene contact. Therefore, providing the model with local context is crucial. At each step of the denoising process, we compute the k-nearest-neighbours (KNN) between the current transformed object point cloud $\hat{\mathbf{O}}_t$ and the scene cloud $\mathbf{S}$.
\item \textit{Translation equivariance.} To allow our model to leverage translation equivariance, we use the centroid $\overline{\mathbf{c}}_t$ of the KNN points $\hat{\mathbf{S}}_t$ to centre both $\hat{\mathbf{S}}_t$ and $\hat{\mathbf{O}}_t$ into a common reference frame with $\overline{\mathbf{S}}_t=\hat{\mathbf{S}}_t-\overline{\mathbf{c}}_t$, $\overline{\mathbf{O}}_t=\hat{\mathbf{O}}_t-\overline{\mathbf{c}}_t$.
\end{itemize}

\begin{figure}[b]
    \centering
    \setlength{\fboxsep}{5pt}%
    \setlength{\fboxrule}{0.5pt}%
    \fbox{\includegraphics[width=1\linewidth-11pt]{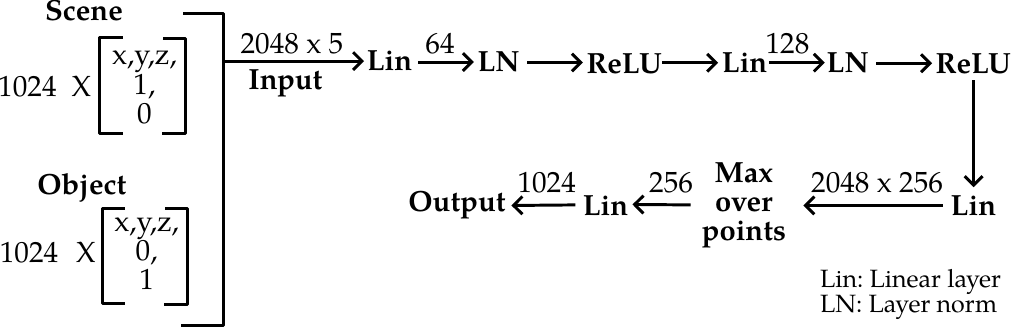}}
    \caption{Point cloud encoder architecture used in this work. The input is made of point position coordinates (x, y, z) and one-hot segmentation labels. The model outputs a 1024-dimensional embedding.}
    \label{fig:point_encoder}
\end{figure}

Finally, the scene and object point clouds are concatenated into point cloud $\overline{\mathbf{P}}_t=(\overline{\mathbf{S}}_t, \overline{\mathbf{O}}_t$) and input to the point cloud encoder architecture $\mathbf{f}_{\boldsymbol\phi}$ described in \cref{fig:point_encoder} to obtain the latent scene representation $\mathbf{y}_t = \mathbf{f}_{\boldsymbol\phi}(\overline{\mathbf{P}}_t)$.

\subsection{Diffusion Model}
\begin{figure*}[t]
    \centering
    \begin{tikzpicture}[
        >=Latex,
        box/.style={draw, rectangle, minimum height=7mm, minimum width=18mm, align=center, font=\small},
        lbl/.style={font=\scriptsize, align=center},
        redarrow/.style={<->, red, thick},
        redloop/.style={->, red, thick},
        redlabel/.style={red, font=\scriptsize},
    ]

     \node[lbl] (scene) at (-0.3,-1.5) {Scene \& Object\\Input};
    \node[box] (scheduler) at (2,0) {DDPM\\Scheduler};
    \node[box] (unet1) at (6,0) {Pose Diffusion\\U-Net};
    \node[lbl] (denoised1) at (8,0) {Denoised\\Pose};
    
    \node[lbl] (time) at (0, -0.5) {Time Step};
    \node[lbl] (stable) at (0, 0) {Stable\\Placement};
    \node[box] (observation) at (2,-1.5) {Observation\\Processing};
    \node[lbl] (noisy) at (4, 0) {Noisy\\Pose};
    \node[box] (encoder) at (6,-1.5) {Point Cloud\\Encoder};
    
    \draw[->] (scene) -- (observation);
    \draw[->] (observation) -- node[midway, lbl, above] {Updated Local\\Context}(encoder);
    \draw[->] (encoder) -- node[midway, lbl, right] {Latent\\Representation} (unet1);
    \draw[->] (unet1) -- (denoised1);
    
    \draw[->] (time) -- (scheduler);
    \draw[->] (stable) -- (scheduler);
    \draw[->] (scheduler) -- (noisy);
    \draw[->] (noisy) -- (unet1);
    
    \draw[redarrow] (denoised1.north) |- ++(0,0.4) -| (stable.north);
    \node[redlabel] at (4,0.9) {Mean Squared Error (MSE)};

    \draw[->, -{Stealth}] (denoised1.south) |- ++(0,-1.9) -| (observation.south);
    
    \draw[dashed] (9.5,0.75) -- (9.5,-3);
    
    \node[box] (unet2) at (11.7,-1.25) {Pose Diffusion\\U-Net};
    \node[lbl] (denoised2) at (14,-1.25) {Denoised\\Poses};
    \node[box] (validity) at (16,-1.25) {Validity\\Check};
    \node[lbl] (stable2) at (16,-2.5) {Stable\\Placements};
    
    \node[lbl] (latent2) at (11.7,0) {Scene \& Object\\Input};
    \node[lbl] (noise) at (11.7,-2.5) {Gaussian\\Noise};

    \node[box] (gradient) at (14, 0) {Stability\\Gradient};
    
    \draw[->] (latent2) -- (unet2);
    \draw[->] (noise) -- (unet2);
    \draw[->] (unet2) -- (denoised2);
    \draw[->] (denoised2) -- (validity);
    \draw[->] (validity) -- (stable2);
    \draw[->] (latent2) -- (gradient);
    \draw[->] (denoised2) -- (gradient)  ;
    \draw[->] (gradient) -- (denoised2) ;
    
    \draw[redloop] (denoised2.south) ++(0,0) arc[start angle=25,end angle=-150,radius=0.6] -- ++(0,0.25) ;
    \node[redlabel] at (13.5,-2.0) {$\times 50$};

    \node at (4,-3.25) {(a) Training};
    \node at (14,-3.25) {(b) Inference};
    
    \end{tikzpicture}
    \vspace{-3mm}
    \caption{Training and inference pipelines of the proposed stable placement pose generative model. (\textbf{Left}) During training, a point cloud encoder maps the local context extracted from input scene and object data to a latent representation, which provides conditioning to a denoising U-Net. 
    (\textbf{Right}) At inference time, the U-Net iteratively denoises a sample, initially set to Gaussian noise, to generate a batch of candidate poses. The stability gradient of poses generated at intermediate denoising steps further guides the inference process. A validity test ensures that only stable placements are retained.}
    \label{fig:architecture}
\end{figure*}
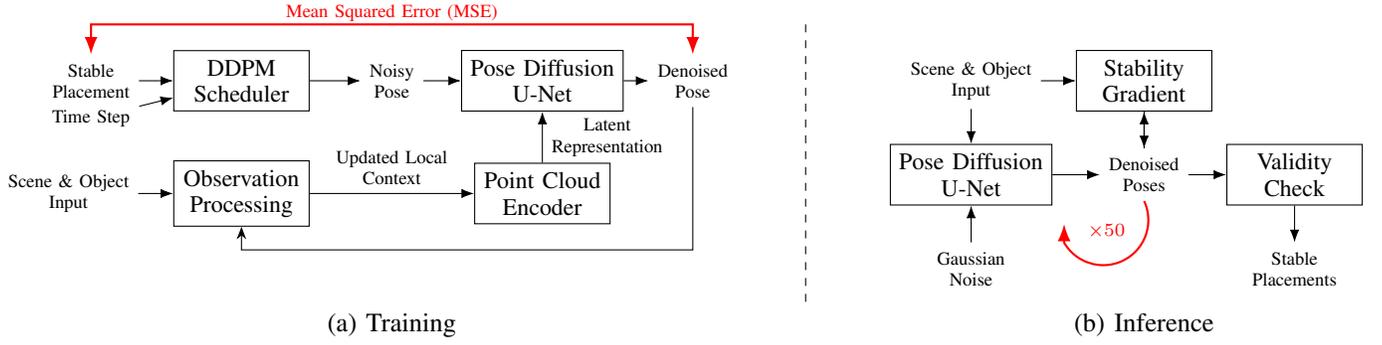
\textit{Training}. We train a U-Net model $\hat{\mathbf{T}}_{\boldsymbol{\theta}}$ \cite{janner_2022_planning}, to predict a placement pose $\mathbf{T}_0$ given a noised version $\mathbf{T}_t$, the current noise time step $t$, and a feature embedding $\mathbf{y}_t$ of the observation $\mathbf{P}$ computed per \cref{sec:obs_processing}. The objective in \cref{eq:loss_function} thus becomes
\begin{align}
    \label{eq:loss_function_pose}
    \mathcal{L}(\mathbf{T}_0, \boldsymbol{\theta})=\sum_{\mathbf{T}_0\in \mathcal{T}} \sum_{t=1}^T \|\hat{\mathbf{T}}_{\boldsymbol{\theta}}(\mathbf{T}_t,t, \mathbf{y}_t)-\mathbf{T}_0\|^2.
\end{align}
Note that $\mathbf{T}_t$ does not necessarily remain in $\rm{SE}(3)$, as Gaussian noise is added to $\mathbf{T}_0$ according to the squared cosine schedule described in  \cite{ho_2020_ddpm}. 

\textit{Stability loss}. In addition to generating poses that have a high likelihood under the training data distribution, we would like to penalize weakly stable placement poses. Consider the differentiable loss function
\begin{equation}
    \label{eq:stability_loss}
    \mathcal{J}(\mathbf{T}) = \frac{-1}{|\mathbf{S}|}\sum_{S_i\in\mathbf{S}}\sum_{O_i\in\mathbf{O}}\lvert\mathbf{n}_{S_i} \cdot \mathbf{n}_{O_i}\rvert \cdotp \exp\{{\frac{\|\mathbf{p}_{S_i}-\mathbf{T}\mathbf{p}_{O_i}\|}{-d_{max}}}\}\cdot \text{R}(S_i),
\end{equation}
where $d_{max}$ controls the steepness of the exponential function.
At a point of contact between two objects, the positions of the scene and object points is almost exactly equal and $\Norm{\mathbf{p}_{S_i}-\mathbf{T}\mathbf{p}_{O_i}}\approx 0$. Thus, minimizing this distance helps to achieve a geometrically precise placement.
Additionally, the surface normals at the scene and object points should be directly opposed and $\lvert\mathbf{n}_{S_i} \cdot \mathbf{n}_{O_i}\rvert \approx 1$ such that contacting objects are not penetrating.
Finally, the odds that a placement will result in a stable scene are maximized when scene points supporting the object are as robust as possible. This is reflected in \cref{eq:stability_loss} by weighting scene points with their robustness values.
Hence, \cref{eq:stability_loss} is minimized when scene points with the highest robustness values are in direct contact with object points having directly opposing normals.
We refer to using our differentiable stability loss to achieve conditional generation as \textit{robustness guidance}.

\textit{Inference}. 
Once the diffusion model $\hat{\mathbf{T}}_{\boldsymbol{\theta}}$ is trained, we sample a stable placement by following the inverse Langevin dynamics sampling process
\begin{align}
    \label{eq:placement_sampling}
    \mathbf{T}_{t-1} = \alpha_t(\mathbf{T}_t + \beta_t\hat{\mathbf{T}}_{\boldsymbol{\theta}}(\mathbf{T}_t,t, \mathbf{y}_t)) - \gamma_t \nabla\mathcal{J}(\mathbf{T}_t)) + \boldsymbol{\xi}_t \
\end{align}
from $t=T$ to $1$, with $\boldsymbol{\xi}_t \sim \mathcal{N}(\mathbf{0};\sigma_t \mathbf{I})$, and $\alpha_t$, $\beta_t$, $\sigma_t$ following a squared cosine scheduling \cite{ho_2020_ddpm}. We apply robustness guidance in \cref{eq:placement_sampling} by computing the negative gradient of the stability loss \cref{eq:stability_loss} with respect to the current noisy pose $\mathbf{T}_t$, and summing the gradient to the denoising process. 

Note that our guidance neither affects the training phase of the diffusion model nor requires training an external model to obtain the guidance gradient; it is applied solely during inference without special adjustments. To further minimize the computational overhead of guidance, we apply a constant $\gamma_t$ at regular intervals of every $n$ time step, and zero it out otherwise. Finally, although all available point features are required for guidance, only the points' positions and one hot encodings are required for neural network training and inference. 

After obtaining the denoised sample $\mathbf{T}_{t-1}$ according to the sampling process in \cref{eq:placement_sampling}, we perform Gram-Schmidt orthogonalization on the last 6 elements of $\mathbf{T}_{t-1}$ to convert the 6D rotation representation to a rotation matrix in $\rm{SO}(3)$ \cite{zhou_2019_continuity}. This new pose, denominated $\hat{\mathbf{T}}_{t-1}$, can then be used to transform the object cloud with respect to the scene to place the object or to change the observation representation $\mathbf{O}$ in order to obtain feature embeddings $\mathbf{y}_{t-1}$ for the next denoising step as described in \cref{sec:obs_processing}. 

\subsection{Performance Metrics}
\label{sec:perf_metrics}
Placements generated by our diffusion model are evaluated for robustness, non-penetration and stability.
We designate $\StableSet$, $\UnstableSet$, $\PenetratingSet$, and $\PenetrationFreeSet$ as the set of stable, unstable, penetrating, and penetration-free placements, respectively.
The set of all generated placements is $\AllPlacementsSet=\StableSet \cup \UnstableSet \cup \PenetratingSet \cup \PenetrationFreeSet$, and $\Abs{\AllPlacementsSet}$ is the cardinality of $\AllPlacementsSet$.
We define the following performance criteria:
\begin{itemize}
    \item Validity rate: Among all placements, proportion of the placements that are valid. Given by $\Abs{\StableSet \cap \PenetrationFreeSet}/\Abs{\AllPlacementsSet}$.
    \item Valid over penetration-free: Among the placements that are not penetrating, the proportion of placements that are also stable, and hence valid. Given by $\Abs{\StableSet \cap \PenetrationFreeSet}/\Abs{\AllPlacementsSet \setminus \PenetratingSet}$.
\end{itemize}
While the validity rate is ultimately what we care about, the valid over penetration-free criterion focuses on stability by disregarding placements that are invalid due to penetrations.

In addition to the criteria defined above, the minimum and median robustness of the resulting scene is computed with \cite{nadeau_2025_robustness} for all valid placements to compare between planners on the ability to generate robust placements.

\section{Experiments}
\label{sec:experiments}
We benchmark methods on the task of stably placing a cube (of homogeneous mass density and whose size is 2 m$^3$) in the four scenes defined in \cref{fig:schematics}. A placement is considered to be valid if it fulfills the criteria defined in \cref{sec:placement_planning_problem_def}. Placement quality is evaluated using the metrics outlined in \cref{sec:perf_metrics}.

To highlight the generalizability of our approach, we train four distinct models independently on different subsets of the data.
For each scene, a model is trained on a subset of the data that does not contain any placement example from that scene. A dataset subset not containing any example from scene \textit{S} is referred to as \textit{S Out}. 
For instance, the \textit{Cantilever Out} model is trained on examples from the \{\textit{Table}, \textit{Shelf}, \textit{Balance}\} data subset.
We perform experiments with our \textit{Cantilever Out}, \textit{Table Out}, \textit{Shelf Out}, and \textit{Balance Out} models to evaluate the ability of the proposed method to quickly generate robust placements in previously unseen scenes.
Finally, to account for randomness, we train each model three times with different seeds and report the average performance. Our models are trained on a dataset of 3,200 stable placement examples across four scenes. Considering random seeds and scene augmentations, we comprehensively evaluate our approach through 4,800 experiments across four models.

When evaluating the capacity of a model to generate stable placements in a scene, we distinguish between \textit{known scenes} (KS), \textit{unknown scenes} (US), and average (ALL) performance. US performance is measured on a scene that was not part of the training dataset while ALL is the average between KS and US performances. 

We train our models for 4,000 epochs with a batch size of 32 using the AdamW optimizer \cite{loshchilov2017AdamW} and a cosine learning rate scheduler. We train our diffusion model for $T=100$ noising steps and use $T=50$ steps during the denoising process to decrease generation time while still obtaining robust placement poses. We set the guidance weight to $\gamma_t=0.1$ and the guidance interval to $n=2$ to allow the diffusion model to guide the generation towards non-penetrating object poses. We did not tune hyperparameters to maximize performance.

\subsection{Placement Validity Rate}
While our approach generates a batch of placements at once, there is no guarantee that any of the placements will be valid (i.e., stable and non-penetrating). Since each placement validation operation increases the total planning time, increasing the odds of sampling a valid placement (the \textit{validity rate}) reduces the expected planning time.
Hence, we perform experiments to evaluate the validity rates of our models with and without robustness guidance.
Each of our \textit{Cantilever Out} (C-O), \textit{Table Out} (T-O), \textit{Shelf Out} (S-O), and \textit{Balance Out} (B-O) models is used to generate 100 placements that are uniformly divided across 10 variations of the US scene obtained with the transformations outlined in \cref{sec:dataset_generation}.
The performance criteria defined in \cref{sec:perf_metrics} are evaluated for each model and the results are recorded in \cref{tab:validity_rates}.
\begin{table}[ht]
\centering
\caption{Placement Validity (Valid) and Valid Over Penetration-free (V/PF) Rates For Our Four Models}
\label{tab:validity_rates}
\begin{tabular}{l|cccc|cccc}
\toprule
\multirow{3}{*}{\textbf{Model}} 
  & \multicolumn{4}{c|}{\textbf{Guidance-free}} 
  & \multicolumn{4}{c}{\textbf{Rob.-guided}} \\
\cmidrule(lr){2-5}\cmidrule(lr){6-9}
  & \multicolumn{2}{c|}{\textbf{Valid \textcolor{red}{$\uparrow$}}} 
  & \multicolumn{2}{c|}{\textbf{V/PF \textcolor{red}{$\uparrow$}}} 
  & \multicolumn{2}{c|}{\textbf{Valid \textcolor{red}{$\uparrow$}}} 
  & \multicolumn{2}{c}{\textbf{V/PF \textcolor{red}{$\uparrow$}}} \\
& \textbf{ALL} & \textbf{US} & \textbf{ALL} & \textbf{US} 
& \textbf{ALL} & \textbf{US} & \textbf{ALL} & \textbf{US} \\
\midrule
C-O & $75$ & $47$ & $84$ & $53$ & $89$ & $84$ & $99$ & $94$ \\
T-O & $66$ & $34$ & $80$ & $48$ & $56$ & $32$ & $86$ & $50$ \\
S-O & $78$ & $92$ & $86$ & $99$ & $71$ & $85$     & $99$ & $100$ \\
B-O & $77$ & $85$ & $85$ & $91$ & $72$ & $94$ & $99$ & $100$ \\
\midrule
Average & $\textbf{74}$ & $65$ & $84$ & $73$ & $72$ & $\textbf{74}$ & $\textbf{96}$ & $\textbf{86}$\\
\bottomrule
\end{tabular}
\end{table}

\subsection{Placement Robustness Assessment}
Placements obtained with our proposed approach making use of robustness guidance are compared to those obtained with the same models not using guidance and to those obtained with the planner defined in \cite{nadeau_2025_planning}.
To assess placement quality, the robustness to normal forces applied at surface points is computed for the scene \textit{after} placing the object in the selected pose.
Minimum and median robustness values are used to quantify placement quality, the average value being always infinite in our scenes.

For each scene, we generate 100 placements using \cite{nadeau_2025_planning}, using our robustness-guided model not trained on the scene (Rob-Guid.), and using the same model with guidance disabled with $\gamma_t=0$ in \cref{eq:langevin_guidance} (No Guid.).
We randomly generate 10 variations of the scene, with each variation obtained by transforming the original scene point cloud according to the steps outlined in \cref{sec:dataset_generation}.
For each scene variation, 10 placements are generated and the median robustness value of the scene is computed for valid placements. The average and standard deviation of the minimum and median robustness values computed across all scene variations are reported in \cref{tab:min_rob} and \cref{tab:med_rob}, respectively.
\begin{table}[ht]
\centering
\caption{Minimum Scene Robustness (N) After Placing}
\label{tab:min_rob}
\begin{tabular}{l|c|c|c}
\toprule
\textbf{}  & \textbf{Using \cite{nadeau_2025_planning}} & \textbf{Ours (No Guid.)} & \textbf{Ours (Rob-Guid.)} \\
\midrule
Cantilever  & $1.81 \pm 0.96$  & $1.55 \pm 1.11$   & $\textbf{3.05} \pm 0.76$\\
Table       & $1.60 \pm 0.68$  & $1.27 \pm 0.68$   & $\textbf{2.45} \pm 0.11$\\
Shelf       & $2.94 \pm 1.17$  & $2.97 \pm 1.09$   & $\textbf{3.71} \pm 0.22$\\
Balance     & $3.28 \pm 0.96$  & $2.56 \pm 1.09$   & $\textbf{3.81} \pm 0.32$\\
\midrule
Average & $2.41$ & $2.09$ & $\textbf{3.26}$ \\
\bottomrule
\end{tabular}
\end{table}

\begin{table}[ht]
\centering
\caption{Median Scene Robustness (N) After Placing}
\label{tab:med_rob}
\begin{tabular}{l|c|c|c}
\toprule
\textbf{}  & \textbf{Using \cite{nadeau_2025_planning}} & \textbf{Ours (No Guid.)} & \textbf{Ours (Rob-Guid.)} \\
\midrule
Cantilever  & $11.61 \pm 0.81$    & $10.47 \pm 2.36$      & $\textbf{12.41} \pm 0.90$  \\
Table       & $5.96 \pm 0.71$     & $5.45 \pm 1.03$       & $\textbf{6.89} \pm 0.12$  \\
Shelf       & $9.67 \pm 0.51$     & $9.77 \pm 0.23$       & $\textbf{9.78} \pm 0.05$  \\
Balance     & $9.15 \pm 0.82$     & $8.33 \pm 1.35$       & $\textbf{9.63} \pm 0.21$ \\
\midrule
Average & $9.10$ & $8.51$ & $\textbf{9.68}$ \\
\bottomrule
\end{tabular}
\end{table}

\subsection{Planning Time Evaluation}
Our approach and \cite{nadeau_2025_planning} are compared in terms of stable placement planning time in \cref{tab:running_time}.
The average time required to produce a placement with our approach is measured by generating a batch of ten placements on an NVIDIA RTX A6000 GPU. The experiment is performed five times and the average inference time span per placement is computed. The time taken to compute scene robustness and to validate placement validity is then added to the inference time span, and the result is recorded in \cref{tab:running_time}.

\begin{table}[ht]
\centering
\caption{Stable Placement Generation Time (s)}
\label{tab:running_time}
\begin{tabular}{l|c|c}
\toprule
\textbf{}  & \textbf{Using \cite{nadeau_2025_planning}} & \textbf{Ours (Rob-Guided)} \\
\midrule
Cantilever  & $0.59 \pm 0.42$     & $\textbf{0.32} \pm 0.09$\\
Table       & $\textbf{0.26} \pm 0.17$     & $0.42 \pm 0.24$\\
Shelf       & $\textbf{0.26} \pm 0.09$     & $0.27 \pm 0.08$\\
Balance     & $0.76 \pm 0.45$     & $\textbf{0.28} \pm 0.08$\\
\midrule
Average & $0.47$ & $\textbf{0.32}$ \\
\bottomrule
\end{tabular}
\end{table}

\section{Discussion}
\label{sec:discussion}
In this work, we propose a diffusion-based approach for generating a variety of stable placements in previously unseen scenes.
Our method makes use of \textit{robustness guidance} at inference time to encourage our model to select placement poses resulting in scenes that can better resist external force perturbations.
In \cref{sec:experiments}, we benchmark our proposed approach (Rob-Guid.) against a guidance free variation (No Guid.) and against the planner proposed in \cite{nadeau_2025_planning}.

\subsection{Results Overview}
\label{sec:results_overview}
As expected, results in \cref{tab:validity_rates} indicate that the valid over penetration-free ratio is higher than the validity ratio, highlighting the critical but difficult problem of planning collision-free placements.
Nonetheless, the validity rates are always greater than 32\% when evaluated on US, on average requiring only about three samples to yield a valid placement.
According to results in \cref{tab:validity_rates}, the validity rate of the guidance-free model on ALL is about 9\% greater than the validity rate measured on US.
Notably, \cref{tab:validity_rates} results' suggest that robustness guidance improves the valid over penetration-free ratio on ALL and US.

\begin{figure}[h]
    \centering
    \subfloat[\textit{Cantilever} without guidance]{
      \setlength{\fboxsep}{0pt}%
      \setlength{\fboxrule}{0.5pt}%
      \fbox{\includegraphics[width=0.48\columnwidth-1pt]{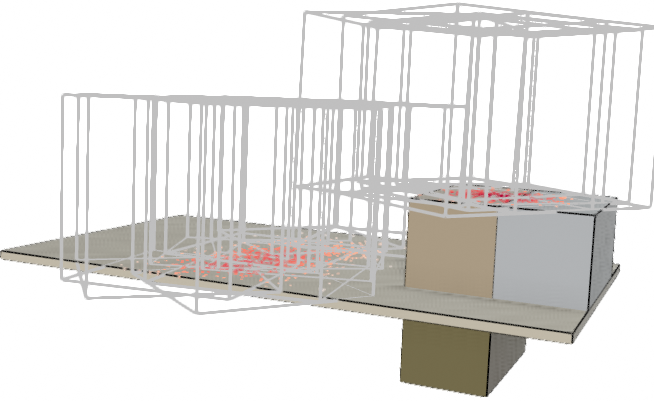}}
    }%
    \subfloat[\textit{Table} without guidance]{
      \setlength{\fboxsep}{0pt}%
      \setlength{\fboxrule}{0.5pt}%
      \fbox{\includegraphics[width=0.48\columnwidth-1pt]{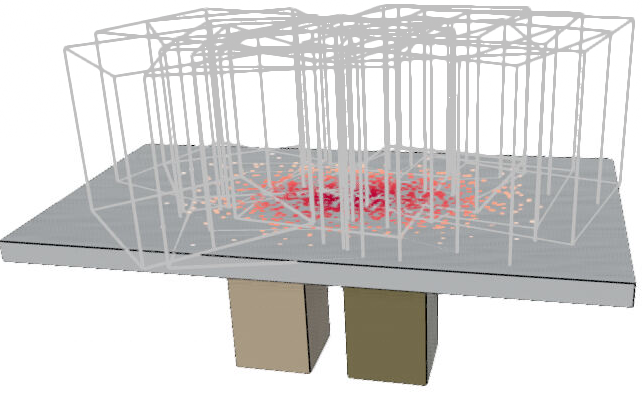}}
    }\\
    \subfloat[\textit{Cantilever} with guidance]{
      \setlength{\fboxsep}{0pt}%
      \setlength{\fboxrule}{0.5pt}%
      \fbox{\includegraphics[width=0.48\columnwidth-1pt]{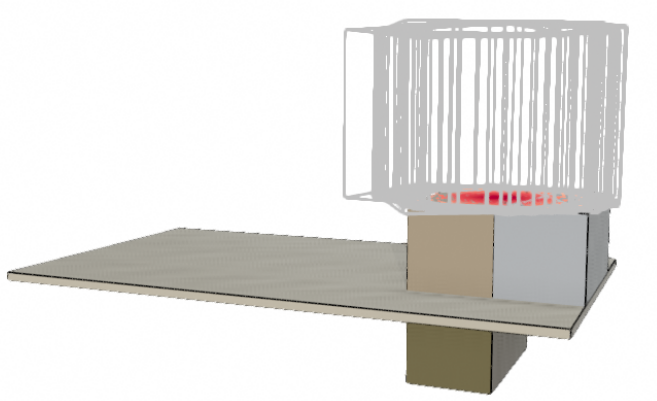}}
    }%
    \subfloat[\textit{Table} with guidance]{
      \setlength{\fboxsep}{0pt}%
      \setlength{\fboxrule}{0.5pt}%
      \fbox{\includegraphics[width=0.48\columnwidth-1pt]{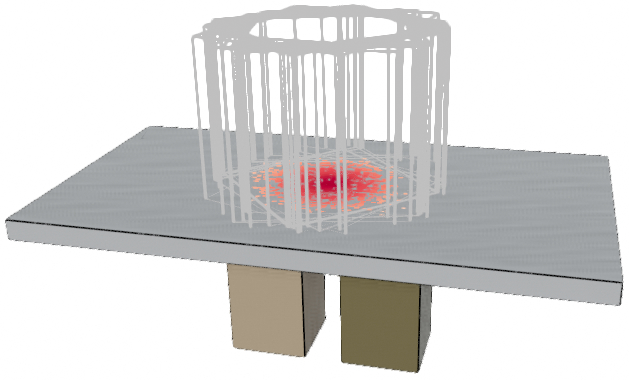}}
    }
    \caption{Placement distribution examples generated with models not trained on the scene shown when robustness guidance is not used (top), and when it is used (bottom). Grey outlines indicate placement poses and red dots indicate contact points.}
    \label{fig:comparison}
\end{figure}

According to \cref{tab:min_rob}, our robustness-guided models produce placements resulting in scenes that are significantly more robust on average. 
The minimal robustness of the scene after placing the object with our models, in previously unseen scenes, is on average 35\% better than \cite{nadeau_2025_planning} and 56\% better than our guidance-free models.
This is visualized in \cref{fig:comparison}, in which several placements generated for the \textit{Cantilever} and \textit{Table} scenes are shown. When robustness guidance is not used, many of the placements generated (i.e., those on the large slab) result in the scene being almost destabilized, and hence yield low minimal scene robustness. In contrast, when robustness guidance is applied, placements are mostly distributed on the top of the cantilever or in the centre of the table, where they can improve the stability of the structure.

Results in \cref{tab:med_rob} show that the median robustness is 6\% better with our robustness-guided models than \cite{nadeau_2025_planning} and 14\% better than guidance-free models.
Additionally, our results suggest that robustness guidance   improves placement consistency; the standard deviation of the median scene robustness is 5.7 smaller than when not using guidance and 5.5 times smaller than when using \cite{nadeau_2025_planning}.
Clearly, guiding the sampling process with robustness information can enable diffusion models to produce more stable placements in unknown scenes.

A critical aspect of practical stable placement planning concerns the time required to determine a valid placement.
Compared to \cite{nadeau_2025_planning}, our approach yields stable placements about 47\% faster on average, according to our results in \cref{tab:running_time}.
While the difference in planning time is negligible for scenes like \textit{Shelf} where stable placement planning is relatively straightforward, this difference becomes noticeable for scenes like \textit{Balance}.
This can be due to \cite{nadeau_2025_planning} being misled into trying to place in some high-robustness areas (e.g., the bottom slab of \textit{Balance}) that result in the object colliding with the scene. 
Our model seems to be less affected by this issue.

\subsection{Limitations}
\label{sec:limitations}
We have yet to investigate the extent to which our proposed approach could be used in scenes where sliding and rolling effects could undermine stability, or when the mass distribution of the object being placed varies. 

Our method represents scene and object geometry using point clouds.
While convenient, point clouds are inherently incapable of describing volumes, and therefore hinder learning about object solidity. This can explain why applying robustness guidance can increase the proportion of penetrating placements in some scenes (e.g., with the \textit{Shelf Out} model).
An improved version of our method could use meshes or geometric primitives to describe the scene, which could more effectively penalize object penetration during training.

While our work assumes object models to be available, which is a reasonable assumption in industrial contexts, our approach could not be directly applied in unknown environments.
In practice, a perception pipeline would precede our framework to provide the latter with object model estimates.

\section{Conclusion}
\label{sec:conclusion}
In this work, we proposed a diffusion-based approach for stable placement planning -- a challenging problem in robotic manipulation.
We extensively evaluated our method in four multi-object scenes requiring reasoning about stability.
Our robustness-guided diffusion model can generate stable placement poses in previously unseen scenes, achieving improved robustness to external forces, and lower running time compared to a state-of-the-art stable placement planner and a guidance-free diffusion model.
Moreover, our stability-aware loss can be directly incorporated into off-the-shelf models without additional training.
We consider our work to be a stepping stone toward developing models capable of reasoning about multi-object statics.
We expect that scaling up our method with a larger dataset would yield a fast and general approach to stable placement planning.

\bibliographystyle{ieeetr}
\bibliography{references}

\end{document}